\documentclass[a4paper]{article}

\usepackage{INTERSPEECH2020}

\usepackage{microtype}
\usepackage{graphicx}
\usepackage{subfigure}
\usepackage{booktabs} 
\usepackage[T1]{fontenc}
\usepackage{amsmath}
\usepackage{multirow}
\usepackage{enumitem}

\title{Speech To Semantics: Improve ASR and NLU Jointly via All-Neural Interfaces}
\name{Milind Rao, Anirudh Raju, Pranav Dheram, Bach Bui, Ariya Rastrow}
\address{
  Amazon Alexa, USA}
\email{\{milinrao, ranirudh, pddheram, bachbui, arastrow\}@amazon.com}

\def \figpath {./}

\begin{document}

\maketitle
\begin{abstract}
We consider the problem of spoken language understanding (SLU) of extracting natural language intents and associated slot arguments or named entities from speech that is primarily directed at voice assistants. Such a system subsumes both automatic speech recognition (ASR) as well as natural language understanding (NLU). An end-to-end joint SLU model can be built to a required specification opening up the opportunity to deploy on hardware constrained scenarios like devices enabling voice assistants to work offline, in a privacy preserving manner, whilst also reducing server costs. 

We first present models that extract utterance intent directly from speech without intermediate text output. We then present a compositional model, which generates the transcript using the Listen Attend Spell ASR system and then extracts interpretation using a neural NLU model. Finally, we contrast these methods to a jointly trained end-to-end joint SLU model, consisting of ASR and NLU subsystems which are connected by a neural network based interface instead of text, that produces transcripts as well as NLU interpretation. We show that the jointly trained model shows improvements to ASR incorporating semantic information from NLU and also improves NLU by exposing it to ASR confusion encoded in the hidden layer. 
\end{abstract}

\section{Introduction}
\label{sec:introduction}

Spoken dialog systems such as those utilized in voice assistants such as Alexa, Siri and Google Home, typically consists of a sequential chain of sub-systems, including Spoken Language Understanding (SLU), Dialog Management, Natural Language Generation and Text-to-Speech. Generally, these sub-systems perform cloud-based processing of speech, following on-device wakeword detection. First, the SLU system extracts natural language semantics such as utterance intent as well as associated named entities or slot values from the speech segment. The appropriate application is then invoked for further execution and finally responses are processed by a text-to-speech system and relayed to the user. An example of intents and slots for an utterance is in Table.\ \ref{tab:utt_eg}. Conventionally, the SLU system comprises two distinct stages: (1) An Automatic Speech Recognition (ASR) system obtains the transcript or a text representation of the raw audio segment, (2) A Natural Language Understanding (NLU) system subsequently consumes the transcript or alternatively n-best hypotheses of the ASR system and extracts semantics, in particular, the domain, intent and slots.

In this work, we consider neural end-to-end (E2E) SLU models that produce semantics of intents and slots from audio. A primary motivation for the work arises from deployment of SLU systems to devices that are more resource limited than cloud servers. For such devices, a neural E2E SLU model can be customized under the given resource constraints to satisfy a limited set of intents or use cases (eg. home automation or auto), and deployed. Moving SLU computation from cloud to devices allows (1) offline use, e.g. in cars or emergency situations (2) latency gains from placing computations closer to the user (3) cost and carbon savings from reduced fleet sizes and reducing communication payloads. 

E2E SLU provides an alternative paradigm to the conventional approach of compressing individual components of the ASR or NLU systems to satisfy on-device resource constraints. In the latter approach, the NLU subsystem is not exposed to audio information such as prosody, or ambiguity in the ASR decoding beyond n-best hypotheses. Errors from the ASR system cascade down to NLU tasks. Our approach to developing E2E SLU systems leverages models developed in ASR and NLU communities by replacing the text interface between them with a neural network hidden interface layer. We term this interpretable subclass of E2E SLU models Joint SLU models that produce intermediate transcript as well as NLU annotations. We show that NLU metrics improve with exposure to this richer interface and also that ASR metrics improve from NLU feedback with joint training. The multitask training of these models can make use of datasets with only transcribed audio as well as audio with NLU annotations.

\begin{table}
\begin{tabular}{| p{0.2\linewidth} | p{0.7\linewidth} |}
\hline
Transcript & set an alarm for six a.m \\ \hline 
Intent & SetNotificationIntent \\ \hline 
Annotation & set|\textbf{Other} an|\textbf{Other} alarm|\textbf{NotificationType} for|\textbf{Other} six|\textbf{Time} a.m.|\textbf{Time} \\ \hline
Slots & NotificationType - alarm, Time - six a.m. \\ \hline
\end{tabular}
\caption{An example of intent, slots for an utterance. \label{tab:utt_eg}}
\vspace{-1cm}
\end{table}

\begin{figure*}[h]
    \includegraphics[width=\linewidth]{\figpath 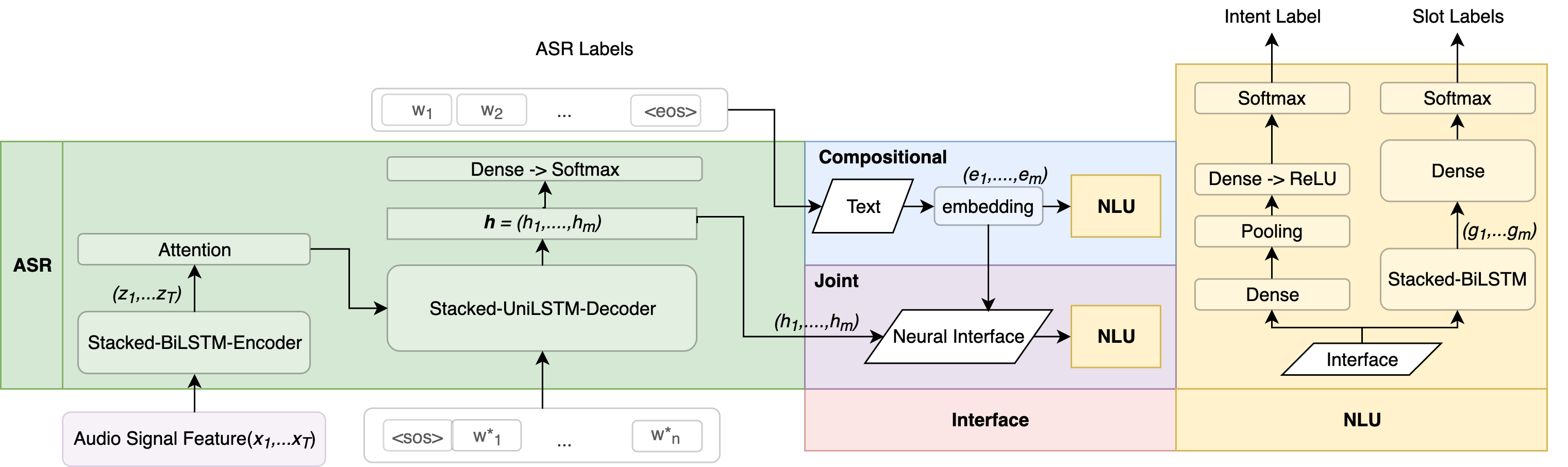}
    \caption{E2E SLU Architectures. Includes ASR subsystem, Neural NLU subsystem, Compositional pipeline and joint pipeline \label{fig:e2e_slu}}
    \vspace{-0.5cm}
\end{figure*}

\subsection{Prior work}

A few prior works have considered the E2E SLU problem. In the work from Google \cite{haghani2018audio}, authors first develop the problem and note that having an intermediate text representation improves performance. They consider encoder-decoder sequence networks to predict transcript and a serialized form of the semantics in a multitask model where decoders are separate, a two-stage model where the transcript is obtained first, and a joint model where a single decoder predicts both jointly. In contrast, we formalize distinctions between ASR and NLU subsystems and study the impact of end-to-end training, and interfaces such as text or hidden layers between the two. In \cite{ghannay2018end}, a CTC based network is used to extract named entities from French speech while we use attention based networks and train on larger corpora. Finally, \cite{serdyuk2018towards, qian2017exploring} are works that obtain a single label (intent or domain) directly from speech segments. Our work goes beyond this and performs slot filling as well. 

E2E SLU models are similar to other multitask speech systems such as speech translation systems or multilingual ASR systems. This work has been enabled by advancements in E2E ASR, where such systems have been shown to outperform conventional RNN-HMM hybrid ASR systems \cite{chiu2018state, prabhavalkar2017comparison} when trained on large acoustic datasets. Connectionist Temporal Classification networks \cite{graves2006connectionist} was the first all neural E2E ASR model that trained a Recurrent Neural Network (RNN) on audio input features with a transcript label sequence of a different length by considering all possible alignments between inputs and labels. In Recurrent Neural Network-Transducer \cite{graves2012sequence}, authors extend CTC by also modelling interdependencies between input-output and output-output distributions using an added prediction network. In both cases, an efficient forward-backward computation enables loss computation and backpropagation over all alignments. In contrast to the aforementioned streaming architectures, in attention based sequence-to-sequence networks such as Listen Attend Spell (LAS) \cite{chan16}, input features are processed by encoder networks that produce a hidden representation output for each feature. The decoder estimates an element of the label sequence at each step using an attention network to focus on a fraction of the encoder network outputs.

Extracting intent and slots from transcript is a long running problem in NLU \cite{lample2016neural, kim2017onenet}. In survey \cite{wu2017clinical}, authors compare DNN and earlier feature engineering approach coupled with conditional random fields or softmax layers for the purpose of named entity recognition. The interface between ASR and NLU systems has traditionally been the best hypothesis sequence although richer interfaces such as lattices and word confusion networks have also been well studied \cite{hakkani2006beyond, henderson2012discriminative, tur2002improving}. In this work, we develop a simple joint intent and slot prediction network and study the impact of text vs hidden layer interfaces between ASR and NLU.

\subsection{Contributions}

In Sec.\ \ref{sec:models}, we first present a low-resource streaming model that extracts utterance intent directly from speech without intermediate text output. We then present a compositional model that is similar to a non-streaming pipelined two-stage SLU architecture, where a LAS based ASR system produces a transcript which is then consumed by an independently trained Neural NLU system. Finally, we present the aforementioned E2E differentiable Joint SLU models where the interface between ASR and NLU is a shared hidden layer. We restrict ourselves to 1-best interfaces between ASR and NLU and leave n-best hidden layer interfaces to future work. 

We present experimental results, baselines, and metrics on a variety of datasets in Sec.\ \ref{sec:evaluation} and answer the following questions:
\begin{itemize}[leftmargin=*] 
\item Does jointly predicting the intermediate ASR output help with the tasks of intent recognition and slot filling?
\item Can we improve performance of a compositional two-stage ASR and NLU system with a richer interface than text?
\item Is system performance in terms of ASR and NLU metrics, improved with joint training?
\end{itemize}

\section{Models}
\label{sec:models}

\subsection{Direct Audio to Intent}
\label{sec:audio_to_intent}

The first class of models corresponds to a sequence classification task where we extract intent directly from speech. Use cases for this model arise in highly resource constrained devices that require models of low footprint or as a short-lister for NLU domains to speed up computation.

Audio features for time instants $t\in \{1,2,\ldots,T\}$ are inputs to a stacked LSTM encoder. Outputs $y_t$ are globally pooled, passed through a 2 layer feed forward Relu network and a dense layer to obtain intent label logits. The model is trained via cross-entropy loss. A unidirectional RNN layer would be appropriate for low latency streaming compatibility compared to a bidirectional model.  

\subsection{Compositional Models}
\label{sec:compositional}

Here, we describe a compositional SLU model where two independently trained models (ASR and NLU) are pipelined with the interface being the 1-best ASR hypothesis. These models contrast with the earlier audio to intent model as it performs slot filling in addition to intent recognition and provides intermediate ASR output. First, we describe an LAS based ASR system, followed by a simple neural NLU model and finally describe how these are combined in the resulting compositional model. 

\textbf{ASR Subsystem - LAS}: The ASR subsystem is an attention based sequence-to-sequence Listen Attend and Spell (LAS) model \cite{chan16}.  Similar to other seq2seq models, LAS primarily comprises 2 components - a stacked RNN encoder which encodes audio frames $(x_1,...,x_T)$ to generate representations $\textbf{z} = (z_{1},...,z_{T^{'}})$, and an auto-regressive RNN decoder which sequentially generates logits or subword probability distribution by attending to \textbf{z}. The set-up is described by the green box in Fig.\ \ref{fig:e2e_slu}. In addition to the set-up above, we improve performance by employing multiple Bahdanau attention heads \cite{bahdanau2016end}, augmenting the audio features via Specaugment \cite{park2019specaugment}, employ a stochastic tokenizer based on a unigram model for subword regularization \cite{kudo2018subword}, and finally do label smoothing on the output logits \cite{chiu2018state}. 

\textbf{Neural NLU}: A neural NLU model in this context accepts the utterance transcript as input and produces both intent as well as associated slots as described in the yellow box of Fig.\ \ref{fig:e2e_slu}. The reference annotations as seen in Tab.\ \ref{tab:utt_eg} consists of words in the transcript tagged with their corresponding slot (which may include the non-informative Other tag). After performing a unigram subword tokenization of the transcript, all constituent subwords of a word are tagged with the latter's slot. During inference, the constituent subwords are combined to form the word and the tag for the last constituent wordpiece is taken as the tag or slot for the word.

Embeddings of transcript subwords $\textbf{e} = (e_{1},...,e_{m})$ are used for two purposes:  (1) Slot filling: the embeddings are passed to a stacked BiLSTM layer. Layer output $\textbf{g} = (g_{1},...,g_{m})$  is passed to a dense layer that maps $g_i$ to slot logits (2) Intent classification: similar to the non-RNN layers in Sec.\ \ref{sec:audio_to_intent}, pooling and dense layers produce intent logits from $\textbf{e}$. The model is trained by backpropagating the sum of the two cross-entropy loss functions.  

\textbf{Composing ASR and NLU} : A first system that accepts speech input and produces transcript, intent, slots can now be obtained by composing the above described ASR and NLU systems. Audio $\rightarrow$ 1-best transcript $\rightarrow$ NLU intent, slots.

\subsection{Joint SLU Models}

In Sec.\ \ref{sec:compositional}, we described a pipelined two-stage model comprised of independently trained ASR and NLU systems. In that model, any ASR error had a downstream impact on NLU results, and ASR training was not cognizant of losses beyond cross-entropy or possibly word error rate measures. In this section, we develop an end-to-end fully differentiable SLU model.

This joint SLU model leverages the LAS ASR models and Neural NLU models developed earlier. The model, as illustrated in Fig.\ \ref{fig:e2e_slu}, consists of:
\begin{itemize}[leftmargin=*]
\item LAS based ASR system: Similar to the ASR sub-system of Sec. \ref{sec:compositional}, at each decoding step $i$, the RNN decoder attends to the encoded representation \textbf{z} and emits an intermediate decoder hidden representation $h_i$ corresponding to the output of the final LSTM layer. This is passed through a dense network to obtain subword logits.
\item Neural NLU: Instead of accepting embeddings $e_{i}$ for decoded subword units at each step $i$ as in the NLU of the compositional model, the input to the NLU system here is the LSTM layer output $h_i$ concatenated with the embeddings  $e_{i}$.  The rest of the Neural NLU model is the same as the architecture described in Sec \ref{sec:compositional}, resulting in intent classification logits for the utterance and slot logits for each subword.
\item ASR-NLU interface: The interface between ASR and NLU is the LSTM output $h_i$ at step $i$ concatenated with the decoded subword embedding $e_i$  i.e.\ $(h_i, e_i)$ is the input to the NLU model.

\end{itemize} 

\section{Experimental Setup}
\label{sec:evaluation}

\subsection{Datasets}

We utilize in-house far-field acoustic datasets that include parallel speech transcripts and NLU annotations of intent and slots. The datasets we report on include (1) 13 intent data: A training set with 1150 hrs of data (1.36 M utterances) filtered across 13 intents of interest such as WhatTime, WhatDay, CancelNotification, Snooze, etc from four domains namely Global, Help, HomeAutomation and Notifications. The chosen intents include 59 slots that do not rely on personalization or large catalogs. (2) More intent data: Unfiltered training data of 6000 hrs (8 M utterances) classified as one of 42 of intents or other, spanning 150 slots (3) Clean dataset of 4M utterances (2.5k hrs) with utterances from 18 intents and 79 slots that have been normalized to standardize human annotations (4) 23khr ASR-only data: This is an dataset of transcribed speech used to pretrain the ASR LAS models.

\subsection{Metrics}

\begin{table*}[t]
\centering
\begin{tabular}{|c|p{4.1cm}|c|*{4}{c|} p{3.4 cm} |}
\hline
&\textbf{Model} & \textbf{Params}(M) & \textbf{WER} & \textbf{SemER} & \textbf{IRER} & \textbf{ICER} & \textbf{Details}  \\ \hline
1&Oracle Neural NLU baseline & - & - & 0.588 &  0.571 & 0.437 & ground truth text input\\ \hline \hline
2& Compositional: LAS $\rightarrow$ text $\rightarrow$ NLU & 81 & 1 & 1 & 1 & 1 & ASR, NLU independently trained \\ 
2a& Compositional LAS* $\rightarrow$ NLU & 81 & 0.938 & 0.989 & 0.990 & 0.972 & ASR further fine tuned \\ \hline
3& SLU Joint LAS $\leftrightarrow$  NLU & 88 & 0.962 & 0.973 & 0.985 & 0.934 & ASR, NLU jointly trained \\ 
3a& SLU Joint N-best Oracle & 88 & 0.365 & 0.557 & 0.603 & 0.377 & Best results from 4 hyps \\ \hline
\end{tabular}
\caption{Comparison of all models on NLU and ASR metrics for the clean 18 intent dataset (eval set size 697k utterances or 406 hrs). 4-beam decoding has been used for compositional and joint models. Numbers relative to compositional model.  \label{tab:crosstown_performance}}
\vspace{-1cm}
\end{table*}

We use a combination of metrics to assess the performance of ASR and NLU systems. 

\noindent\textbf{Intent Classification}: Intent Classification Error Rate (ICER) is the primary metric for evaluating intent. This is a recall based metric. 

\noindent\textbf{Speech Recognition}: Word Error Rate (WER) computed as the normalized ratio of edit distance or Levenshtein distance to sequence length. Edit distance is calculated as the length of the shortest sequence of insert, delete, and substitute operations (over words) to transform one sentence to another.

\noindent\textbf{Slot Filling}: The Semantic Error Rate (SemER) metric is used to evaluate jointly the intent and slot-filling performance or NLU performance. Comparing a reference of words and their accompanying tags, performance is classified as: (1) Correct slots - Slot name and slot value correctly identified, (2) Deletion errors - Slot name present in reference but not hypothesis, (3) Insertion errors - Extraneous slot names included by hypothesis, (4) Substitution errors - Slot name from hypothesis is included but incorrect slot value. Intent classification errors are substitution errors.
\begin{align}
\text{SemER} &= \frac{\text{\#Deletion} + \text{\#Insertion} + \text{\#Substitution}}{\underbrace{\text{\#Correct} + \text{\#Deletion} + \text{\#Substitution}}_{\text{\#Slots in Reference}}}
\end{align}
The Interpretation Error Rate (IRER) metric is related and is the ratio of utterances for which a semantic error has been made. 



\subsection{Model details}
\textbf{Audio features}: audio encoder inputs are global mean and variance normalized 64-dim Log Filter Bank Energy (LFBE) coefficients computed on a 25 ms window with 10 ms shifts. They are downsampled to a 30ms frame rate by stacking the current frame with 3 frames to the left. 

\noindent\textbf{Text features}: the transcript is tokenized to 4500 subword units using unigram language model \cite{kudo2018sentencepiece} and 256 dim.\ embeddings are used. 

\noindent\textbf{Audio encoder}: the LAS model encoder has 5 layer BiLSTM with 512 units (28M params). 

\noindent\textbf{ASR decoder}: LAS decoder has 2 layer LSTM with 1024 units. 4 Bahdanau attention heads with depth 256 and output dimension 768 are concatenated.  LAS system comprised above audio encoder and this decoder. During inference, 4-beam decoding is done and teacher forcing for training. 

\noindent\textbf{NLU decoder}: 2 layer BiLSTM with 256 units. Outputs $\rightarrow$ dense layer to get slot logits. 

\noindent\textbf{Intent classification layer}: inputs (audio encoder//SLU Joint hidden interface) passed through 512 unit dense layer, pooled ,2 512 unit ReLu feed-forward layers. 

\noindent\textbf{Direct audio to intent}: large model has audio encoder feeding into intent classification layer above. A small model has 2x256 LSTM audio encoder feeding to intent classification layer with 256 units. 

\noindent\textbf{Neural NLU}: NLU decoder above consumes subword embedding to produce slots logits as does the intent classification layer to obtain intent classification logits. 

\noindent\textbf{SLU Joint}: LAS attention decoder output $h_i$ concatenated with decoded subword embedding $e_i$ are inputs to the intent classification layer and the neural decoder for slots. The SLU Joint model is trained in a multistep fashion: (1) ASR training: ASR layers loaded from pretrained LAS model trained on 23k hr data and finetuned on the dataset of interest. (2) NLU training: ASR layers marked as non-trainable and NLU layers trained with intent, slot losses; NLU exposed to encoded word confusion through $\mathbf{(h, e)}$ (3) Joint training: the entire network is fine-tuned with a sum of subword, intent, slot loss and ASR weights are also updated from backpropagated gradients of NLU losses. 

\noindent\textbf{Training}:  Adam optimizer was used with learning rate $10^{-5}$ for LAS, SLU Joint, $10^{-4}$ for Neural NLU, $10^{-3}$ for direct audio to intent. LAS training had newbob learning rate scheduling. 4-16 GPUs were used for model training with batch sizes ranging from 64-256 depending on model size. Models were trained for 6 epochs unless noted otherwise to keep number of data points seen consistent and the best models from here were chosen. 

\section{Results}

\noindent\textbf{Joint transcript and intent classification}

From Tab.\ \ref{tab:13_intent_performance}, a larger direct audio-to-intent model brings a $16\%$  improvement over the low footprint model on the 13 intent dataset. Jointly predicting transcript either via compositional or joint models improve intent classification significantly. 

\noindent\textbf{NLU performance metrics}

From Tab.\ \ref{tab:crosstown_performance}, we see from rows 1, 2 that NLU metrics degrade on neural NLU models on ASR 1-best hypothesis instead of the ground truth. In row 2a, we use a stronger LAS trained using minimum word error rate and NLU metrics further improve on compositional models. In row 3, the joint ASR-NLU model with the neural network interface of text embeddings and the hidden layer from decoder RNN is pretrained from the LAS model of row 2. The NLU metrics improve by 2.7\% on SemER, 1.5\% on IRER, and 3.6\% on intent classification.The improvement to NLU metrics from joint training is also seen in Tab.\ \ref{tab:13_intent_performance} with a marginal improvement from the hidden interface and training only the NLU subsystem and a larger improvement from joint training. Error bars on ICER are $\pm 2.3\%$, $\pm 0.86\%$ on SemER, and $\pm 1.04\%$ on IRER. 
  
\noindent\textbf{ASR performance metrics}

In Tab.\ \ref{tab:crosstown_performance}, we compare rows 2 and 3 to see joint training with NLU feedback improving WER by 3.8\% and a larger 6.6\% in Tab.\ \ref{tab:13_intent_performance}. Contrasting rows 2a, 3 in Tab.\ \ref{tab:crosstown_performance}, even if we improve ASR through external methods such as sequence loss training, better ASR performance does not translate to better performance on NLU metrics suggesting that joint training improves ASR performance on words that impact downstream NLU performance. In row 3a, we see 4-best oracle performance compares with the oracle neural NLU performance of row 1 with a large reduction in ASR and NLU metrics. Error bars are $\pm 1.54\%$ on WER. 

\begin{table}\centering
\begin{tabular}{| p{3.6cm} | c | c | c|}
\hline
\textbf{Model} & \textbf{WER} & \textbf{SemER} & \textbf{ICER} \\ \hline
Oracle Neural NLU & - & 0.879 &  0.719  \\ \hline
Audio-intent 1.M params low footprint & - & - & 1.792 \\
Audio-intent 29M params & - & - & 1.496 \\ \hline
Compositional & 1 & 1 &  1 \\ \hline
SLU Joint ASR $\rightarrow$ NLU& 1 & 0.918 &  0.906  \\ 
SLU Joint ASR $\leftrightarrow$ NLU& 0.934  & 0.883 & 0.967 \\ \hline
\end{tabular}
\caption{Model performance on the 13 intent dataset with 4-beam decoding, eval set size 135k, hidden+embedding interface. All results are relative to the compositional model and models have been trained for 3 epochs. \label{tab:13_intent_performance}}
\vspace{-1 cm}
\end{table}

\noindent\textbf{Dealing with Out of Domain Data}

For the 13 intent dataset, we use the larger More intent dataset to train an out of domain detector. The SLU Joint model is trained on this dataset with 42 intents + other. If the intent is classified by this model to be any of the 29 intents (or other) that do not overlap with the 13 intents, it is classified as out of domain. Thus, we have two intent classifiers running: one that classifies if it is one of 13 intents and the other to classify if it is one of 42 intents + other. The latter is used to identify OOD utterances and if it is not, the former classifies it. This system can be tuned to achieve a desired false accept rate.

\section{Conclusion}

We developed models for the problem of spoken language understanding of extracting natural language intent and named-entities or slots directly from speech. Such an end-to-end model can be customized and deployed on resource constrained device enabling new offline and privacy focussed use cases. We first developed an audio to intent model of small footprint. We then developed a compositional model with a pretrained LAS ASR model whose outputs, the transcription of the audio, is fed to a pre-trained neural NLU model. Finally, an end-to-end fully differentiable, interpretable Joint SLU model was presented where the NLU system consumes not the transcript output of the LAS system but a neural network interface that encodes ASR word confusion. These models were trained on multiple datasets and that affirmed the following points: (1) intent classification performance improves when ASR outputs are also produced (2) replacing a text or wordpiece interface between compositional ASR and NLU systems with a neural network hidden and joint training leads to a 2.7+\% relative improvement to NLU metrics for intents, and slots and mitigates the downstream impact of ASR errors (3) joint training reduces ASR WER by 3.8\% through backpropagation of NLU losses to the ASR layers. 

\vspace{-2 mm}
\section{Acknowledgments}

The authors would like to thank Samridhi Choudhary, Kai Wei, Kanthashree Sathyendra, Joe McKenna, Zhe Zhang, Jing Liu for support on the data preparation pipeline, Gautam Tiwari for helpful discussions and Shehzad Mevawalla, Jagannath Krishnan and Bjorn Hoffmeister from the management team for their support.

\bibliographystyle{IEEEtran}

\bibliography{e2eslu_interspeech}

\end{document}